\documentclass{article}

\usepackage[preprint]{neurips_2026}
\usepackage{amssymb} 
\usepackage[utf8]{inputenc}
\usepackage[T1]{fontenc}
\usepackage{hyperref}
\usepackage{url}
\usepackage{booktabs}
\usepackage{amsmath}
\usepackage{amsfonts}
\usepackage{nicefrac}
\usepackage{microtype}
\usepackage{xcolor}
\usepackage{graphicx}
\usepackage{algorithm}
\usepackage{algpseudocode}
\usepackage{fvextra}
\fvset{fontsize=\footnotesize,breaklines=true,breakanywhere=true}

\title{KACE: Knowledge-Adaptive Context Engineering for Mathematical Reasoning}

\author{%
  Jayant Parashar \\
  School of Computing \\
  University of Georgia \\
  \texttt{Jayant.Parashar@uga.edu} \\
  \And
  Suchendra M.\ Bhandarkar \\
  School of Computing \\
  University of Georgia \\
  \texttt{suchi@uga.edu} \\
}

\hypersetup{pdftitle={KACE: Knowledge-Adaptive Context Engineering for Mathematical Reasoning},pdfauthor={Jayant Parashar, Suchendra M. Bhandarkar}}

\begin{document}

\maketitle

\begin{abstract}
Context-engineering methods improve a frozen language model by learning text from execution feedback and placing it in the prompt. When one learned prompt carries everything that was learned, irrelevant lessons compete with relevant ones, and the system can keep only what fits in that prompt. We present Knowledge-Adaptive Context Engineering (\textsc{KACE}), which decides per problem both how much inference to spend and which learned knowledge to show. A fixed tiered self-consistency schedule (2, 3, then 5 samples behind agreement gates) lets easy problems exit after two agreeing samples, and the tier at which a problem exits serves as its difficulty label. Offline, a teacher-guided learner distills failed training traces into short typed cards kept in a store partitioned by difficulty tier and domain, and each escalated tier reads only the cards of its (tier, domain) cell. On MATH-HARD, \textsc{KACE} reaches $77.3\%$ against $72.7\%$ for Best-of-$5$ self-consistency with fewer solver calls, and it scores above tiered ACE ($70.0\%$) and tiered GEPA ($72.0\%$). It also achieves the highest accuracy on AIME 2025 ($61.1\%$ against $53.3\%$ for Best-of-$5$ and $57.8\%$ for tiered GEPA) and on OlymMATH-EN ($48.1\%$). A component analysis shows that the store adds $4.0$ points on MATH-HARD and $3.7$ on OlymMATH-EN over the tiered schedule alone, and that the (tier, domain) reading rule outperforms every alternative rule for choosing cards from the same store.
\end{abstract}

\section{Introduction}

Context engineering improves frozen language models by learning the information placed around a problem, leaving the weights unchanged. Methods such as ACE, which grows an evolving playbook of lessons \citep{zhang2025ace}, and GEPA, which evolves a prompt by reflective search \citep{agrawal2025gepa}, learn from execution feedback, and in both the learned artifact is also exactly what the solver reads. For mathematical reasoning this coupling has two costs. Lessons that are irrelevant to the current problem sit in the prompt next to the useful ones, and everything the system keeps must fit in one active context.

A second line of work spends inference compute unevenly across problems. Self-consistency samples several solutions and votes \citep{wang2022selfconsistency}, and adaptive or cascade methods stop early on problems that look easy \citep{aggarwal2023adaptive,snell2024computeoptimal,kim2024mdagents,chang2026cascadedebate}. These methods decide how much to sample, but every sample sees the same context.

\textsc{KACE} uses one signal for both decisions. A fixed tiered self-consistency schedule samples two attempts that must agree, then three, then five, then pools all ten; the tier at which a problem exits is a cheap, model-derived difficulty label. Offline, a teacher model reads the solver's failed training traces and writes short typed cards (a lemma, an invariant, a decomposition, a verification check), each tagged with the domain of the failure it came from and a difficulty tier. At test time, problems that exit at the first tier read no cards, and each escalated tier reads only the cards filed under its own tier and the problem's domain. The learned store can therefore hold more than any single solver call reads, and easy problems never pay for it.

\textsc{KACE} ranks first on all three benchmarks we evaluate, against Best-of-$5$ self-consistency and against ACE and GEPA run on the same tiered schedule. On MATH-HARD, the largest test split, it gains $4.7$ points over Best-of-$5$ with about a third fewer solver calls. Removing one component at a time shows that both contribute: the tiered schedule drives the gain on AIME 2025, the card store adds $4.0$ points on MATH-HARD and $3.7$ on OlymMATH-EN, and among the card-selection rules we tried, the fixed (tier, domain) rule is the one that adds accuracy.

\paragraph{Contributions.}
\textbf{(1)} \textsc{KACE}, a context-engineering method that keys learned context on the exit tier of adaptive self-consistency: a partitioned card store read one (tier, domain) cell at a time, built by a two-phase offline learner that adds cards from cell-local failures and then edits them under the updated store (§\ref{sec:method}). \textbf{(2)} Results on AIME 2025, MATH-HARD and OlymMATH-EN, where \textsc{KACE} scores above Best-of-$5$, tiered ACE and tiered GEPA on the same tiered schedule, with per-problem paired tests and a token and offline-budget accounting (§\ref{sec:main-exp}, Appendices~\ref{app:budget-comparison} and~\ref{app:paired-stats}). \textbf{(3)} A component analysis separating the tiered schedule from the store, and an ablation of the card-selection rule over a fixed store (§\ref{sec:components}, §\ref{sec:ablations}).

\section{Related Work}
\label{sec:related}

\paragraph{Learned context from feedback.} ACE accumulates a playbook of lessons from execution feedback \citep{zhang2025ace}, GEPA evolves a prompt by reflective mutation under a validation filter \citep{agrawal2025gepa}, and DSPy-style optimizers tune instructions and demonstrations \citep{khattab2024dspy,opsahlong2024mipro}. MCE casts context engineering as bi-level optimization, in which a meta-level agent evolves the skills that a base-level agent uses to build context artifacts \citep{ye2026mce}. Each produces one artifact that every call reads. \textsc{KACE} learns a store of cards and reads a different cell of it per tier and domain.

\paragraph{Memory and retrieval of reusable reasoning.} Dynamic Cheatsheet keeps an adaptive memory across test problems \citep{suzgun2025dynamic}. Buffer of Thoughts distills reusable thought templates and retrieves one per problem \citep{yang2024bot}; Didolkar et al.\ label math problems with skills and retrieve skill-matched exemplars \citep{didolkar2024metacognitive}. Agent memories such as ExpeL, Agent Workflow Memory, ReasoningBank, A-Mem and Mem0 store insights or workflows extracted from past episodes \citep{zhao2024expel,wang2024awm,ouyang2026reasoningbank,xu2025amem,chhikara2025mem0}. These systems retrieve by similarity, skill or task. \textsc{KACE} retrieves by a fixed key, the tier at which self-consistency escalated plus a domain label, and our ablation compares that key against similarity and LLM selection over the same store (§\ref{sec:ablations}).

\paragraph{Adaptive test-time compute.} Adaptive-Consistency stops sampling once votes agree \citep{aggarwal2023adaptive}, compute-optimal scaling allocates test-time compute by difficulty \citep{snell2024computeoptimal}, and Adaptive-RAG routes queries by estimated complexity \citep{jeong2024adaptiverag}. The tiered schedule in \textsc{KACE} is a fixed instance of early-exit self-consistency; what is new is using its exit tier to choose which learned context each tier reads. Appendix~\ref{app:related} gives an extended discussion.

\section{Method}
\label{sec:method}

\subsection{A Partitioned Card Store}
\label{sec:formalism}

\textsc{KACE} wraps a frozen solver with an external store $\mathcal{K}$ of typed cards. Let $\mathcal{T}=\{\mathrm{ES},\mathrm{MS},\mathrm{HS}\}$ be the easy, medium and hard solver tiers and $\mathcal{D}$ a set of domains. The store is partitioned into cells
\begin{equation}
    \mathcal{K}=\{\mathcal{K}_{b}\}_{b\in\mathcal{B}},
    \qquad
    \mathcal{B}=\mathcal{T}\times\mathcal{D},
\end{equation}
plus a set $\mathcal{K}_{\mathrm{univ}}$ of universal cards. Cells do not inherit from one another; the only structural question is which cell a call reads. For a problem $x$ the active context is
\begin{equation}
    C(x;\theta_R,\mathcal{K})
    =
    \mathrm{Concat}\big(x,\rho(x;\theta_R,\mathcal{K})\big),
\end{equation}
where
\begin{equation}
    \begin{split}
    &\rho(x;\theta_R,\mathcal{K})
    = \\
    &\begin{cases}
    \emptyset, & \theta_R(x) \in \{\mathrm{ES}\}\times\mathcal{D},\\
    \mathcal{K}_{\theta_R(x)} \cup \mathcal{K}_{\mathrm{univ}}, & \theta_R(x) \in \{\mathrm{MS},\mathrm{HS}\}\times\mathcal{D}.
    \end{cases}
    \end{split}
\end{equation}
The projection $\theta_R$ maps $x$ to a cell from two signals, the tier reached by tiered self-consistency and a coarse domain label. ES reads no cards, so problems that exit there pay nothing for the store.

\paragraph{Relation to prior context methods.} ACE, GEPA and flat retrieval are restrictions of this context function. ACE collapses both axes into one growing prompt that every call reads; GEPA further compresses the learned object into a single natural-language artifact; flat retrieval keeps an unpartitioned $\mathcal{K}$ and replaces the projection $\theta_R$ with top-$k$ similarity search. \textsc{KACE} keeps the difficulty and domain axes and makes the test-time projection explicit, and §\ref{sec:ablations} compares it with the single-context and similarity-retrieval restrictions over the same store.

\begin{figure}[t]
    \centering
    \includegraphics[width=\textwidth]{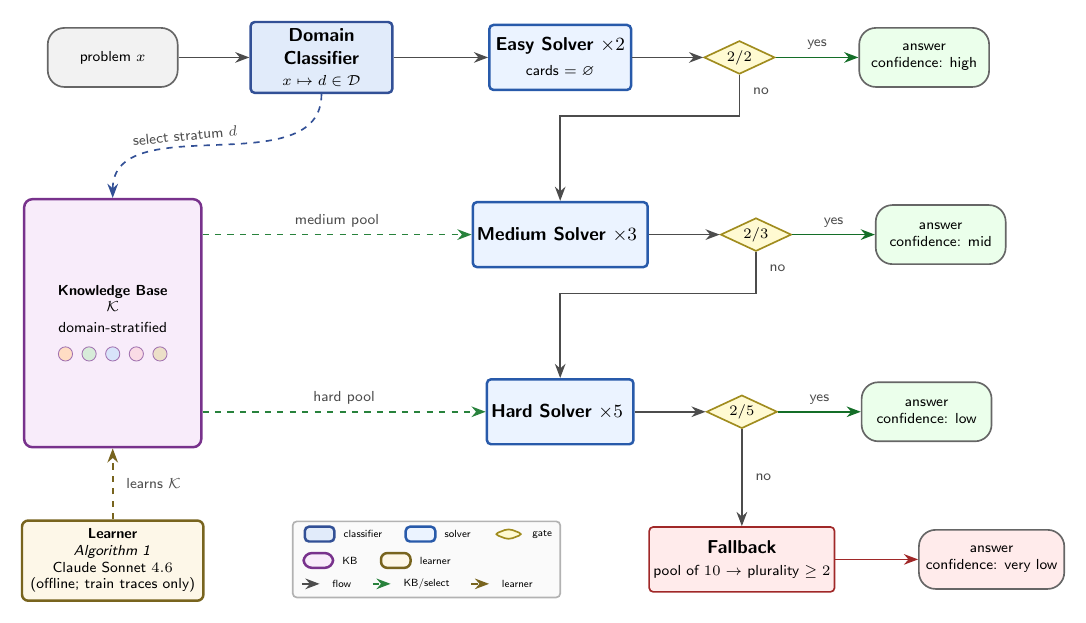}
    \caption{\textsc{KACE} at test time. A frozen LLM is the solver. Tiered self-consistency supplies the difficulty tier and a domain classifier supplies the domain; together they select one (tier, domain) cell of the card store $\mathcal{K}$ (drawn as the ``knowledge base''). ES reads no cards; MS and HS read only their cell plus universal cards. The schedule is fixed: ES ($2$ attempts, $2/2$ gate) $\to$ MS ($3$ attempts, $2/3$ gate) $\to$ HS ($5$ attempts, $2/5$ gate), then a plurality over all $10$ attempts.}
    \label{fig:kace-arch}
\end{figure}

\subsection{Tiered Self-Consistency and Card Reading}
\label{sec:tiered}

A coarse classifier first assigns a domain $d$. ES then samples two attempts and exits if both agree; MS samples three and exits on a $2/3$ majority; HS samples five and exits on a $2/5$ plurality. If no gate fires, the system returns the plurality of all ten attempts when one answer appears at least twice, and otherwise the last HS attempt (Table~\ref{tab:tiered}). The tier at which a gate fires is the difficulty signal used by $\theta_R$: MS reads $\mathcal{K}_{(\mathrm{MS},d)}$ and HS reads $\mathcal{K}_{(\mathrm{HS},d)}$, each with the universal cards and at most 30 cards per attempt. The schedule is the same in every experiment; the main results vary only the store $\mathcal{K}$, and §\ref{sec:ablations} varies the reading rule. In pilot runs, allowing cards at ES increased early exits but also increased confident wrong exits, so ES stays card-free.

\begin{table}[t]
    \centering
    \small
    \begin{tabular}{lllll}
        \toprule
        Tier & Solvers & Cards per attempt & Diversity & Exit gate \\
        \midrule
        ES (Easy Solver)   & 2 & no cards & $2$-lens rotation & $2/2$ unanimous \\
        MS (Medium Solver) & 3 & medium cell at domain $d$ & $3$-lens rotation & $2/3$ majority \\
        HS (Hard Solver)   & 5 & hard cell at domain $d$ & sampling only & $2/5$ plurality \\
        Fallback & \multicolumn{4}{l}{pool $10$ attempts ($2$ ES $+$ $3$ MS $+$ $5$ HS) $\to$ plurality $\ge 2$; else last HS attempt} \\
        \bottomrule
    \end{tabular}
    \caption{Tiered self-consistency in \textsc{KACE}. The schedule, gates and fallback are the same in every experiment. Card visibility is strict: ES $\to$ none, MS $\to$ \{medium, universal\}, HS $\to$ \{hard, universal\}, intersected with the domain.}
    \label{tab:tiered}
\end{table}

\subsection{Offline Construction}
\label{sec:loop}

The learner builds $\mathcal{K}$ from a training split $T=\{(x_i,y_i)\}$ in epochs of two phases. It seeks to reduce the empirical answer loss
\begin{equation}
    \mathcal{L}(\mathcal{K})
    =
    \frac{1}{|T|}\sum_{(x_i,y_i)\in T}
    \ell\!\left(y_i,\pi_S(C(x_i;\theta_R,\mathcal{K}))\right),
\end{equation}
where $\pi_S$ is the frozen solver and $\ell$ is answer-level error. Since $\pi_S$, self-consistency voting and card selection are black-box and non-smooth, we do not compute gradients. Each failed trace instead supplies a natural-language residual: a local description of the knowledge missing in the cell that was actually read.

\paragraph{Phase 1: ADD, cell-local learning from failures.}
At epoch $n$ the tiered solver runs with the current store $\mathcal{K}^{(n)}$, which assigns each training problem a cell $b_i^{(n)}=\theta_R(x_i;\mathcal{K}^{(n)})$ and gives each cell $b$ a residual set
\begin{equation}
    \begin{split}
    R_b^{(n)}
    = {}&\{\,\tau_i^{(n)} : b_i^{(n)}=b,\\
    &\;\; \pi_S(C(x_i;\theta_R,\mathcal{K}^{(n)}))\neq y_i\,\}.
    \end{split}
\end{equation}
ADD updates each cell using only its own residuals:
\begin{equation}
    \widetilde{\mathcal{K}}_{b}^{(n+1)}
    =
    \mathcal{K}_{b}^{(n)}
    \cup
    A_b\big(R_b^{(n)},\mathcal{K}_{b}^{(n)}\big),
\end{equation}
where $A_b$ is a teacher-reflection operator: a teacher model reads the failed traces of that domain and proposes cards, tagging each as medium or hard, for the missing lemma, reduction or check, and a gate-judge screens each card for answer leakage and benchmark-specific wording. This is a cell-local analogue of textual-gradient and reflective prompt updates \citep{pryzant2023protegi,agrawal2025gepa,zhang2025ace}. Because each edit touches only a conditionally read cell, the learner can add many cards in one epoch without rewriting a single shared prompt, as in block-coordinate minimization, which optimizes pieces of a large object from local information while the decomposition is held fixed \citep{razaviyayn2013block}.

\paragraph{Phase 2: REFINE under the updated store.}
The provisional store $\widetilde{\mathcal{K}}^{(n+1)}$ changes the context the solver sees, and with it which tiers fire, which cards are read and which failures remain; a card that helped under $\mathcal{K}^{(n)}$ can become redundant or harmful. The learner therefore reruns the solver,
\begin{equation}
    \widetilde{\tau}_i^{(n+1)}
    \sim
    \pi_S\big(C(x_i;\theta_R,\widetilde{\mathcal{K}}^{(n+1)})\big),
\end{equation}
collects post-update residual sets $\{\widetilde{R}_b^{(n+1)}\}_{b\in\mathcal{B}}$, measures each card's paired lift, and applies a curator
\begin{equation}
    \mathcal{K}^{(n+1)}
    =
    Q\big(
    \widetilde{\mathcal{K}}^{(n+1)},
    \{\widetilde{R}_b^{(n+1)}\}_{b\in\mathcal{B}}
    \big)
\end{equation}
that may keep, edit or deprecate cards but adds none; a rejected edit gets a bounded number of retries $M\!\in\!\{1,2\}$. This pass addresses the induced-distribution problem formalized by dataset aggregation in imitation learning \citep{ross2011dagger}: corrections are evaluated under the distribution that the current policy, here the frozen solver wrapped by the current context function, actually induces. After a small fixed number of epochs the store is frozen before validation and test. Algorithm~\ref{alg:kace} (Appendix~\ref{app:algorithm}) gives the full pseudocode, Appendix~\ref{app:kb-schema} the card schema, and Appendix~\ref{app:example-cards} two cards from the AIME store.

\section{Experiments}

\subsection{Setup}
\label{sec:setup}

\textbf{AIME 2025.} The 30 problems of the 2025 contest form the test set; 90 problems from earlier AIME contests form the training and validation splits (45/45). \textbf{MATH-HARD (L4--L5).} From the Level-4 and Level-5 problems of the MATH dataset of 12{,}500 competition problems \citep{hendrycks2021math}, sampled across its seven subjects, we draw 50 training, 50 validation and 50 test problems. \textbf{OlymMATH-EN (verifiable subset).} From the English statements of the 200 Olympiad-level problems in OlymMATH \citep{sun2025olymmath} we keep the 97 whose final answer is a single closed-form expression that can be checked automatically, split 30/40/27 into train, validation and test.

\paragraph{Solvers and stores.} AIME 2025 and OlymMATH-EN use \texttt{gpt-4.1-mini}; MATH-HARD uses \texttt{gpt-4o-mini}. Claude Sonnet 4.6 is the teacher. Construction runs for two epochs on AIME 2025 and OlymMATH-EN and four on MATH-HARD, chosen on validation; the AIME store has 132 cards (about $33{,}600$ tokens). The offline loop reads only training splits, validation selects hyper-parameters, and test numbers are averaged over $3$ solver-sampling seeds over the frozen store.

\paragraph{Baselines.} \emph{Best-of-$5$ SC} samples five answers with a minimal two-sentence prompt and takes the plurality. \emph{Tiered $+$ ACE} adds a one-epoch ACE playbook and \emph{Tiered $+$ GEPA} a two-epoch GEPA-optimized prompt to the same tiered schedule \textsc{KACE} uses; both artifacts are applied at HS, where they help most, and both use Claude Sonnet 4.6 as reflector. ACE uses one epoch because, in our single-call diagnostic, a playbook grown for two or more epochs falls below the unaugmented solver (Appendix~\ref{app:diagnostic}). \emph{Best-of-$5$ $+$ ACE} adds the AIME playbook to Best-of-$5$. AIME arms marked $^\dagger$ were re-run with per-problem outcomes for the paired tests of Appendix~\ref{app:paired-stats}: a paired bootstrap over problems and an exact sign-flip test on per-problem differences of seed-mean accuracy.

\subsection{Main Results}
\label{sec:main-exp}

\begin{table}[t]
    \centering
    \small
    \begin{tabular}{lccc}
        \toprule
        Method & AIME25 & MATH-HARD & OlymMATH-EN \\
        \midrule
        Solver model & gpt-4.1-mini & gpt-4o-mini & gpt-4.1-mini \\
        \midrule
        Best-of-$5$ SC                        & 53.3\%$^\dagger$ & 72.7\% & 40.7\% \\
        Best-of-$5$ $+$ ACE (minimal prompt)  & 54.4\%$^\dagger$ & --- & --- \\
        Tiered $+$ ACE                        & 54.4\%           & 70.0\% & 40.7\% \\
        Tiered $+$ GEPA                       & 57.8\%$^\dagger$ & 72.0\% & 44.4\% \\
        \textbf{\textsc{KACE}}                & \textbf{61.1\%}$^\dagger$ & \textbf{77.3\%} & \textbf{48.1\%} \\
        \midrule
        \multicolumn{4}{l}{\emph{Component of \textsc{KACE}}} \\
        \textsc{KACE} without the store (tiered SC only) & 60.0\%$^\dagger$ & 73.3\% & 44.4\% \\
        \bottomrule
    \end{tabular}
    \caption{Test accuracy, averaged over three solver-sampling seeds. AIME25 entries marked $^\dagger$ are fully paired re-runs with per-problem outcomes (Appendix~\ref{app:paired-stats}); the Tiered $+$ ACE AIME entry is the original run. The last row removes the card store from \textsc{KACE} and keeps its tiered schedule (§\ref{sec:components}).}
    \label{tab:main-exp}
\end{table}

\textsc{KACE} scores highest on every benchmark (Table~\ref{tab:main-exp}). On MATH-HARD, the largest test split, it reaches $77.3\%$, $4.7$ points above Best-of-$5$ (better on 6 problems, worse on 1) and $5.3$ and $7.3$ points above tiered GEPA and tiered ACE, while averaging $3.41$ solver calls per problem against Best-of-$5$'s $5$ (Appendix Table~\ref{tab:calls-per-problem}). On AIME 2025 it reaches $61.1\%$, $7.8$ points above Best-of-$5$ and $3.3$ above the strongest learned-context baseline, tiered GEPA, at $5.24$ calls per problem; Appendix~\ref{app:paired-stats} gives the per-problem paired tests. On OlymMATH-EN it reaches $48.1\%$ against $44.4\%$ for tiered GEPA. Both single-artifact baselines score at or below the tiered schedule without learned context on all three benchmarks, consistent with the context-bloat failure of a single shared artifact; the ACE playbook added to plain Best-of-$5$ moves it by one problem-run on AIME ($54.4\%$ against $53.3\%$). The same failure appears with a single solver call: on AIME 2025, an ACE playbook grown for two or more epochs scores below the unaugmented solver ($38.9\%$ against $42.2\%$), and splitting the playbook by domain recovers $52.2\%$ (Appendix~\ref{app:diagnostic}).

Each escalated MS or HS attempt on AIME reads about $5{,}560$ learned tokens, a sixth of the $33{,}600$-token store; Appendix~\ref{app:budget-comparison} gives the full token and offline-budget accounting.

\subsection{Where the Gain Comes From}
\label{sec:components}
\label{sec:tiered-vs-bo5}
\label{sec:tier-classifier}

Removing the store and keeping the tiered schedule separates the two components (last row of Table~\ref{tab:main-exp}). The schedule alone matches or exceeds Best-of-$5$ on all three benchmarks at a comparable number of solver calls (Table~\ref{tab:tiered-vs-bo5}), the pattern Adaptive-Consistency reports for early-exit sampling \citep{aggarwal2023adaptive}. The store then adds $4.0$ points on MATH-HARD ($73.3\%$ to $77.3\%$) and $3.7$ on OlymMATH-EN ($44.4\%$ to $48.1\%$). On AIME 2025 the schedule drives the gain, and the store adds a further $1.1$ points.

The exits also provide the difficulty label that selects a cell. Ranking problems by mean exit tier agrees with ranking them by the model's own solve rate on $78\%$ of AIME 2025 pairs and $80\%$ of MATH-HARD pairs, against $50\%$ for a random order (Table~\ref{tab:tier-classifier}). Appendix~\ref{app:tier-classifier} gives the per-bucket breakdown.

\begin{table}[t]
    \centering
    \small
    \setlength{\tabcolsep}{3pt}

\begin{tabular}{llcc}
    \toprule
    Group & Dataset & Best-of-$5$ SC & Tiered SC only \\
    \midrule
    Accuracy & AIME25       & 53.3\% & 60.0\% \\
             & MATH-HARD    & 72.7\% & 73.3\% \\
             & OlymMATH-EN  & 40.7\% & 44.4\% \\
    \midrule
    Calls    & AIME25       & 5.00 & 5.40 \\
             & MATH-HARD    & 5.00 & 3.19 \\
             & OlymMATH-EN  & 5.00 & 5.24 \\
    \bottomrule
\end{tabular}
    \caption{The tiered schedule of \textsc{KACE} without the store, against Best-of-$5$. Calls are average solver calls per problem, computed from the per-tier exit fractions (ES exits at 2 calls, MS at 5, HS and fallback at 10).}
    \label{tab:tiered-vs-bo5}
\end{table}

\begin{table}[t]
    \centering
    \small

    \begin{tabular}{lccc}
        \toprule
        Dataset & $n$ & Spearman $\rho$ & Pairwise concordance \\
        \midrule
        AIME 2025  & 30 & 0.733 & 78\% (302/386) \\
        MATH-HARD  & 50 & 0.816 & 80\% (743/923) \\
        \bottomrule
    \end{tabular}
    \caption{Exit tier against empirical solve rate. Concordance is the fraction of problem pairs whose exit-tier order matches their solve-rate order; $50\%$ is random. Per-bucket breakdowns are in Appendix~\ref{app:tier-classifier}.}
    \label{tab:tier-classifier}
\end{table}

\section{Ablations: How Cards Are Chosen}
\label{sec:ablations}

\begin{table}[t]
    \centering
    \small
    \begin{tabular}{llc}
        \toprule
        Variant & What changes & AIME25 \\
        \midrule
        Full \textsc{KACE} & (tier, domain) cell; fixed reading rule & \textbf{61.1\%} \\
        w/o store & tiered schedule only & 60.0\% \\
        w/o domain axis & same cards, grouped only by tier & 54.4\% \\
        w/o difficulty axis & same cards, grouped only by domain (MS and HS share a cell) & 57.8\% \\
        w/o teacher distillation & raw solver reflections replace distilled cards & 58.9\% \\
        Monolithic context & both axes flattened into one active context & 51.7\% \\
        LLM router & same store, an LLM router replaces the fixed reading rule & 56.7\% (51/90) \\
        Embedding retriever & same store, embedding retrieval replaces the fixed reading rule & 56.2\% (50/89) \\
        \bottomrule
    \end{tabular}
    \caption{AIME 2025 ablations with the tiered schedule fixed; every card-reading variant uses the same 132-card store. The router and retriever rows are selection-rule controls over \textsc{KACE}'s own store, not external memory systems. The monolithic and embedding-retriever rows pool $89$ problem-runs because one problem is missing from one run.}
    \label{tab:ablations}
\end{table}

With the store and schedule fixed, we vary only how cards are chosen (Table~\ref{tab:ablations}). Every alternative selection rule scores below the fixed (tier, domain) rule, by $3.3$ to $9.4$ points, and below the schedule without any store. The largest drops come from removing the domain axis ($54.4\%$) and from showing the whole store at once ($51.7\%$), which is the single-artifact failure again: cards help when the prompt receives the relevant cell and hurt when it receives the rest. Distilled cards beat raw solver reflections by $2.2$ points. An LLM router and an embedding retriever over the same store land within half a point of each other ($56.7\%$ and $56.2\%$), $4.4$ and $4.9$ points below the fixed rule, so the gap does not depend on one routing implementation.

Keeping the domain partition but letting MS and HS read one pooled cell gives $57.8\%$ against $61.1\%$. The pooled cells ($32$--$55$ cards per domain) reach the $30$-card cap, while full \textsc{KACE}'s cells ($20$--$33$) mostly stay below it, so the difficulty split also keeps each prompt focused.

\section{Discussion}

The results support three points. Keying learned context on the exit tier of adaptive self-consistency gives the best accuracy we measured on all three benchmarks, and on MATH-HARD it does so with fewer solver calls than Best-of-$5$. The two components contribute differently: the tiered schedule carries most of the gain on AIME 2025, and the store carries most of it on MATH-HARD and about half on OlymMATH-EN. Learned context needs a narrow reading rule: single shared artifacts fall at or below the tiered schedule without learned context, and every alternative selection rule over our own store falls below it.

The partition lets the store grow without growing the prompt: each escalated attempt reads one cell. The natural next step is a store larger than the solver's context, evaluated at a fixed active-token budget; at that scale we expect the fixed rule to be combined with an embedding retriever within each cell.

\section{Conclusion}

\textsc{KACE} keys learned mathematical knowledge on the difficulty tier that adaptive self-consistency reveals, storing typed cards in a store partitioned by tier and domain and reading one cell per escalated tier. It scores above Best-of-$5$, tiered ACE and tiered GEPA on AIME 2025, MATH-HARD and OlymMATH-EN on the same tiered schedule. Its tiered schedule and its store contribute differently across benchmarks, and the fixed (tier, domain) rule is the only card-selection rule we tested that adds accuracy over the schedule alone.

\section{Limitations}

\begin{enumerate}
    \item \textbf{Small test sets.} The test splits have 30, 50 and 27 problems, so one problem moves a single run by 2--4 points. Paired tests at these sizes (Appendix~\ref{app:paired-stats}) have limited power, so we report the accuracy ordering; larger splits of MATH and OlymMATH would resolve differences of a few points.
    \item \textbf{Baseline differences.} Best-of-$5$ uses a minimal prompt, while the tiered schedule rotates prompt lenses, so the schedule's gain over Best-of-$5$ combines adaptive stopping with prompt diversity \citep{hu2024dipper}. ACE received a smaller offline budget than GEPA and \textsc{KACE} (45 against about 280 and 360 problem-runs), and both single-artifact baselines are applied at HS only.
    \item \textbf{One store per benchmark.} Test seeds resample the solver over one frozen store, so the variance of store construction is not measured. The tier-exit concordance is computed from the same runs it describes, and agreement on a wrong answer can trigger an early exit (Appendix~\ref{app:tier-classifier}).
    \item \textbf{Cost.} \textsc{KACE} moves cost into offline construction and uses about $13\times$ the test-time input tokens of its schedule alone on AIME, so matched solver calls do not imply matched tokens or offline budgets (Appendix~\ref{app:budget-comparison}).
    \item \textbf{Scope.} Stores are learned per benchmark from labeled in-distribution traces, and AIME cards do not help on OlymMATH-EN. All solvers are small OpenAI models with a stronger Anthropic teacher, so part of any gain may be teacher knowledge; we test no open-weight or reasoning-tuned solver.
    \item \textbf{Growing the store.} Context methods help when errors repeat; they do not replace the search that RL or pre-training performs. Two attempts to grow the store beyond observed failures, a synthesis step over disagreeing HS attempts and an ``imagination'' loop that expanded the AIME store from 132 to 450 cards, did not improve test accuracy (Appendices~\ref{app:synthesis} and~\ref{app:imagination}).
\end{enumerate}

\section{Ethics Statement}

\textsc{KACE} stores what it learns as natural-language cards filed by tier and domain, so a person can read, audit and remove any card. The main deployment risk is the failure mode in Appendix~\ref{app:tier-classifier}: the tiered gate can exit early on an answer that all attempts agree on and that is wrong, and the system then reports it with high nominal confidence. Any use of such a system to check student work or scientific calculations needs an independent verifier. The experiments use public benchmarks and hosted models; offline learner and solver costs are reported in Appendix~\ref{app:impl}.

\bibliographystyle{plainnat}
\bibliography{references}

\appendix

\section{Extended Related Work}
\label{app:related}
\label{app:related-work}

\subsection{ACE, GEPA, and MCE}

ACE frames inference-time improvement as context engineering through evolving playbooks \citep{zhang2025ace}. Its strength is that it learns reusable context from experience, but this same accumulation can become a liability on hard math benchmarks if irrelevant guidance grows faster than useful guidance. \textsc{KACE} addresses this limitation by replacing one growing playbook with a structured store of typed cards stratified by difficulty and domain.

GEPA-style systems are closely related in spirit because they also improve prompts or reasoning behavior through iterative feedback and guided refinement. The main difference is that \textsc{KACE} is not optimizing a single prompt artifact; it learns a structured store of reusable knowledge cards and consumes that store under an empirical difficulty signal at test-time. This makes the learning target richer than prompt improvement alone, and lets the stored knowledge exceed what one active prompt holds.

MCE formalizes context engineering as a bi-level optimization: a meta-level agent evolves context-engineering skills, and a base-level agent executes them to optimize context artifacts from training rollouts, with an online setting that continues learning over the test stream \citep{ye2026mce}. \textsc{KACE} occupies a deliberately simpler design point. Its context-engineering operators (propose, curate) are fixed, and there is no outer loop that evolves them. Instead, the card store is constructed once on a training split, scored by paired train-verify lift, and frozen at evaluation; only the test-time projection is dynamic per problem. We also draw a sharper line between \emph{skills} (how to reason or intervene) and \emph{knowledge} (what should be surfaced for a problem family): \textsc{KACE} optimizes the latter exclusively. The empirical bet is that a well-structured, frozen store of typed cards can already capture the bulk of the gains attributed to richer skill-evolution machinery, at substantially lower system complexity.

\subsection{Reflection, Memory, and Self-Improvement}

Reflexion, Self-Refine, and related self-improvement systems show that language feedback can serve as an optimization signal at inference time \citep{shinn2023reflexion,madaan2023selfrefine}. Dynamic Cheatsheet further shows that persistent memory can help difficult reasoning by reusing validated insights across tasks \citep{suzgun2025dynamic}. A more recent line argues that \emph{what stays in working context} is itself a learnable decision: Memory-as-Action \citep{zhang2025memaction} treats context curation as an explicit policy, A-MEM \citep{xu2025amem} and Mem0 \citep{chhikara2025mem0} build production-oriented agentic memory stacks, and MemGPT \citep{packer2023memgpt} formalizes context management as an OS-style problem. \emph{Lost in the Middle} \citep{liu2024lostmiddle} establishes empirically that long context is not free, which motivates keeping the active prompt small. \textsc{KACE} builds on these ideas but makes two changes. First, it converts raw feedback into typed knowledge cards keyed by failure mode and placed into a (difficulty, domain) cell rather than concatenated into one stream. Second, it consults the cards \emph{conditionally} on the empirical difficulty signal at test-time, by exiting card-free at ES and reading only the matching branch after escalation, rather than applying memory uniformly.

\subsection{Prompt Evolution and Modular LM Programs}

A separate family of methods compresses experience into a compact natural-language artifact through reflective or evolutionary search. GEPA \citep{agrawal2025gepa} is our primary foil: its compact-prompt compression is real and powerful, but bounded by the size of a single prompt. DSPy and MIPRO \citep{khattab2024dspy,opsahlong2024mipro} optimize instructions and demonstrations for multi-stage LM programs; ProTeGi \citep{pryzant2023protegi} performs gradient-style prompt search; OPRO \citep{yang2024opro} treats LLMs as optimizers; Promptbreeder \citep{fernando2023promptbreeder} evolves prompts via self-referential mutation; PromptAgent \citep{wang2024promptagent} introduces planning-style prompt search; and SAMMO \citep{schnabel2024sammo} performs symbolic prompt-program search. Each of these methods defines a unit of optimization --- an instruction, a demonstration, a program structure, a search store. \textsc{KACE} extends the list with a different unit: a typed knowledge card that lives at a cell of a card store and does not need to live in any single active prompt.

\subsection{Adaptive Retrieval and Learned Source Selection}

A third line of work treats retrieval, retrieval source, or model choice as a learned action. Self-RAG \citep{asai2024selfrag} casts retrieval as a reflection token, deciding per step whether to retrieve. Adaptive-RAG \citep{jeong2024adaptiverag} routes queries by question complexity to retrieval pipelines of different cost. RouterRetriever \citep{lee2024routerretriever}, Self-Routing RAG \citep{wu2025selfroutingrag}, RAGRouter \citep{zhang2025ragrouter}, Corrective RAG \citep{yan2024crag}, and SeaKR \citep{yao2025seakr} each route among retrievers, sources, or repair operators. RouteLLM \citep{ong2024routellm} routes among models. Two further design points are particularly aligned with \textsc{KACE}: Repoformer \citep{wu2024repoformer} self-supervises selective retrieval by \emph{whether retrieval improves the downstream model}, which is exactly the labeling philosophy behind our paired train-verify helpfulness scores; and FILCO \citep{wang2023filco} learns to filter retrieved context by utility, with an explicit ``show nothing'' action, of which our strict non-cumulative per-tier visibility (hard cards never reach MS) is a static, schema-level analog. \textsc{KACE}'s nearest neighbor among these methods is Adaptive-RAG, which routes by question complexity and is analogous to our tier gate. The novelty is to make the \emph{difficulty signal itself} the test-time projection that selects which slice of a learned, structured knowledge store is read --- with the difficulty signal supplied by per-tier agreement gates rather than by an external classifier.

\subsection{Self-Consistency, Search, Verification, and Compute-Optimal Scaling}

Chain-of-thought prompting and self-consistency established the value of sampled reasoning traces for mathematical reasoning \citep{wei2022cot,wang2022selfconsistency}. DIVERSE \citep{li2023diverse} and DIPPER \citep{hu2024dipper} make diversity and verification independent axes; Tree of Thoughts \citep{yao2023tot} and Graph of Thoughts \citep{besta2024got} generalize sampling into search. A separate verifier line---GSM8K verifiers \citep{cobbe2021gsm8k}, step-level supervision \citep{lightman2023verifystep}, Math-Shepherd \citep{wang2024mathshepherd}, programs-as-verifiers \citep{toh2024programsverifiers}---establishes that selection is often more bottlenecked than coverage. Compute-optimal scaling \citep{snell2024computeoptimal} shows that the right test-time policy depends on problem difficulty, motivating tier-conditioned compute. Adaptive-Consistency \citep{aggarwal2023adaptive} is the closest direct prior on \emph{difficulty-conditioned self-consistency}: it dynamically chooses the per-question sampling budget via an early-exit rule, foreshadowing per-tier exit gates. MDAgents \citep{kim2024mdagents} and CascadeDebate \citep{chang2026cascadedebate} take the same difficulty-routed-compute idea outside self-consistency proper: MDAgents switches between solo and group LLM collaboration by task complexity, and CascadeDebate alternates single-model inference with multi-agent deliberation at escalation boundaries across an LLM cascade. \textsc{KACE}'s tiered self-consistency shape (§\ref{sec:tiered}) sits in the intersection of these lines, with two distinguishing ingredients: (i) strict, non-overlapping per-tier exit gates with fixed attempt counts ($2/3/5$) rather than a continuous early-exit threshold, and (ii) per-tier \emph{context} conditioning (different cell of the card store at each tier), so the effective inference policy is a function of difficulty in both budget \emph{and} prompt.

\subsection{Teacher Distillation and Synthetic Math}

STaR \citep{zelikman2022star}, MetaMath \citep{yu2023metamath}, Math-Shepherd \citep{wang2024mathshepherd}, and ReST-EM \citep{singh2023restem} distill teacher rationales or synthetic data into model weights. \textsc{KACE} leaves weights frozen and distills into a structured external store instead; we cite this line only to disambiguate, not to compete on equal terms.

\subsection{Agentic Search and Workflow Optimization}

Agentic search methods such as AFlow and ADAS search over workflows, decompositions, or system architectures instead of over static prompts \citep{zhang2024aflow,hu2024adas}. These methods show that the structure of inference itself can be optimized. \textsc{KACE} shares this spirit but shifts the search target: the central object is not the workflow but a structured library of typed knowledge cards organized along difficulty and domain.

\subsection{Conditional Computation and Mixtures of Experts}

Routing methods such as RouteLLM dynamically allocate different models or compute levels depending on expected difficulty and cost \citep{ong2024routellm}. Mixture-of-experts architectures generalize the same principle inside the model: only a subset of experts should be active for a given input. \textsc{KACE} extends this conditional-computation perspective to external context. Instead of conditioning only among models or subnetworks, it conditions among knowledge interventions: the empirical difficulty tier and the inferred domain together pick the branch of the card store that is read into the active prompt. That is the key conceptual bridge between conditional computation and context engineering in our framework.

\subsection{Math Benchmarks}

We evaluate on AIME 2025 (Mathematical Association of America), the Hendrycks MATH dataset \citep{hendrycks2021math}, OlymMATH \citep{sun2025olymmath}, and cite OlympiadBench \citep{he2024olympiadbench} and Omni-MATH \citep{gao2024omnimath} for context on olympiad-level evaluation.

\section{Offline Learner Pseudocode}
\label{app:algorithm}

Algorithm~\ref{alg:kace} gives the full pseudocode of the \textsc{KACE} offline learner referenced in §\ref{sec:loop}: a per-cell sweep over learned-card (difficulty, domain) cells in Phase~1, followed by a per-domain revise-and-verify pass with bounded \textsc{Refine} budget in Phase~2.

\begin{algorithm}[t]
\caption{\textsc{KACE Offline Learner} --- block-local ADD followed by induced-distribution REFINE.}
\label{alg:kace}
\begin{algorithmic}[1]
\Require training split $T$; lift threshold $\tau_{\text{lift}}$; epoch budget $N$; refine budget $M \in \{1, 2\}$.
\Ensure frozen card store $\mathcal{K}$ over (tier, domain) cells.
\State $\mathcal{D} \gets \textsc{BuildDomainPartition}(T)$ \Comment{agent call: emits domain partition $\{T_d\}_{d\in\mathcal{D}}$ from a filesystem-level scan of $T$}
\State $\mathcal{K} \gets \emptyset$
\For{$n = 1, \dots, N$}
    \State \textbf{// ===== Phase 1 --- populate the store (add new cards) =====}
    \State \textbf{// 1a: per-cell knowledge sweep over learned-card (difficulty, domain) cells}
    \For{each tier $t \in \{\text{MS, HS}\}$ \textbf{and} each domain $d \in \mathcal{D}$}
        \State $\tau_{t,d} \gets \textsc{TrainRun}(T_d, \mathcal{K})|_t$ \Comment{tier-$t$ slice of the tiered run on $T_d$}
        \State $\mathcal{C}_{t,d} \gets \textsc{FindKnowledge}(\tau_{t,d}, t, d)$ \Comment{agent: propose cards from failures at domain $d$; tier tag set by the teacher}
        \State $\mathcal{K} \gets \textsc{Consolidate}(\mathcal{K} \cup \textsc{GateJudge}(\mathcal{C}_{t,d}))$
    \EndFor
    \State \textbf{// ===== Phase 2 --- revise existing cards via Refine $+$ verify =====}
    \For{each domain $d \in \mathcal{D}$}
        \State $\tau_d' \gets \textsc{TrainRun}(T_d, \mathcal{K})$ \Comment{independent run with the post-Phase-1 store}
        \State $h_d \gets \textsc{Impact}(\tau_d', \mathcal{K}|_d)$ \Comment{paired leave-one-out lift per card at cell $(*, d)$}
        \State $\mathcal{R}_d \gets \textsc{ReflectAndReviseCards}(\mathcal{K}|_d, \tau_d', h_d)$ \Comment{revise existing cards: keep, edit or deprecate}
        \State $\mathcal{K}_{\text{cand}} \gets \textsc{ApplyRevisions}(\mathcal{K}, \mathcal{R}_d)$
        \For{$m = 1, \dots, M$}
            \State $\tau_d^{(m)} \gets \textsc{TrainRun}(T_d, \mathcal{K}_{\text{cand}})$ \Comment{verification rerun}
            \If{$\textsc{Lift}(\tau_d^{(m)}, \tau_d') \ge \tau_{\text{lift}}$}
                \State $\mathcal{K} \gets \mathcal{K}_{\text{cand}}$; \textbf{break} \Comment{commit revision}
            \Else
                \State $\mathcal{K}_{\text{cand}} \gets \textsc{Refine}(\mathcal{K}_{\text{cand}}, \tau_d^{(m)}, \mathcal{R}_d)$ \Comment{agent self-corrects the over-large edit}
            \EndIf
        \EndFor
    \EndFor
\EndFor
\State \textbf{freeze} $\mathcal{K}$
\State \Return $\mathcal{K}$
\end{algorithmic}
\end{algorithm}

\section{Knowledge-Base Schema}
\label{app:kb-schema}

This appendix gives the full card schema used in the card store $\mathcal{K}$, as it appears in the implementation. The schema is intentionally lean: each card stores a content body (the on-disk field name in our implementation logs is \texttt{payload}), a small list of \texttt{routing\_conditions}, and tier and domain tags that determine the card's cell in the store.

\paragraph{Card fields.}
\begin{itemize}
    \item \texttt{card\_id} (\texttt{str}, unique).
    \item \texttt{payload} (\texttt{str}): the natural-language content of the card, shown to the solver when the active cell of the store is read; for medium cards this is typically one identity plus one sanity-check, for hard cards a $4$--$8$-step rubric. The payload is acquired during the offline construction loop and may be edited during REFINE in the current or future epochs.
    \item \texttt{routing\_conditions} (\texttt{list[str]}): a small list of natural-language clauses (selection criteria and exclusion clauses) attached to the card on disk. At runtime, these clauses are surfaced in the card context so the model can choose better from the eligible tier-domain pool; cell position determines eligibility, and routing conditions help select and apply cards within that pool.
    \item \texttt{difficulty\_tag} $\in \{\text{medium}, \text{hard}, \text{universal}\}$: one of the two coordinates of the card's cell in the store; drives strict per-tier visibility (see below). ES is intentionally card-free.
    \item \texttt{domain\_tags} (\texttt{list[str]}): the other coordinate of the card's cell in the store; multi-label, intersected with the active domain at evaluation. The literal tag \texttt{universal} passes any domain.
    \item \texttt{helpfulness\_score} (\texttt{float}): per-card score updated from paired train-verify lift measurements during offline curation.
    \item \texttt{provenance}:
    \begin{itemize}
        \item \texttt{source} (\texttt{str}): e.g., \texttt{teacher\_distillation} or \texttt{winning\_trace:<pid>}.
        \item \texttt{supporting\_problems} (\texttt{list[str]}): training problems on which lift was measured.
        \item \texttt{validated\_lift} (\texttt{str}): a free-form lift annotation, e.g.\ \path{"+2/6 -> +4/6 on P026 (k=6)"}.
        \item \texttt{promotion\_status} $\in \{\text{experimental}, \text{validated}, \text{deprecated}\}$.
        \item \texttt{n\_uses}, \texttt{n\_wins}, \texttt{n\_losses} (\texttt{int}): usage and paired-impact counters.
        \item \texttt{epoch\_introduced} (\texttt{int}): curation-loop epoch in which the card first entered $\mathcal{K}$.
    \end{itemize}
\end{itemize}

\paragraph{Strict per-tier visibility.}
The pipeline reads cards by the (tier, \texttt{difficulty\_tag}) pair using a strict, non-cumulative table:
\begin{center}
\small
\begin{tabular}{lll}
    \toprule
    Tier & Visible difficulty tags & Used by \\
    \midrule
    ES & $\varnothing$          & easy solver \\
    MS & \{medium, universal\}  & medium solver \\
    HS & \{hard, universal\}    & hard solver \\
    \bottomrule
\end{tabular}
\end{center}

We chose non-cumulative visibility after pilot runs in which hard cards were also visible to medium solvers and learned cards were allowed at ES. Strict visibility keeps each cell of the store coherent in difficulty register and avoids leaking hard rubrics into MS, where they were observed to over-anchor solver attempts. ES remains card-free because ES cards increased early exits but also increased wrong early exits.

\paragraph{JSON form.}
Each card is stored in a thread-safe JSON-backed store. A serialized card has the form:
\begin{Verbatim}
{
  "card_id": "...",
  "payload": "...",
  "routing_conditions": ["...", "..."],
  "difficulty_tag": "medium" | "hard" | "universal",
  "domain_tags": ["..."],
  "helpfulness_score": 0.0,
  "provenance": {
    "source": "...",
    "supporting_problems": ["..."],
    "validated_lift": "...",
    "promotion_status": "experimental" | "validated" | "deprecated",
    "n_uses": 0,
    "n_wins": 0,
    "n_losses": 0,
    "epoch_introduced": -1
  }
}
\end{Verbatim}

\section{Example Learned Cards}
\label{app:example-cards}

Cards are distilled exclusively from training-split traces (§\ref{sec:setup}); test-split problems were never inspected during card construction. To make the schema concrete, we show two cards distilled by \textsc{KACE} during training on AIME-style training problems and frozen for evaluation. The first targets a hard-geometry problem family; the second targets a hard-combinatorics family. Both cards were promoted to \texttt{validated} status after positive paired train-verify lift, and both live at the (hard, geometry) and (hard, combinatorics) cells of the store respectively. The fields shown below are the literal contents of the card record (the on-disk field names are preserved verbatim), flattened into the display read into the active prompt when the matching cell is consulted.

\subsection*{Card 1: \texttt{ALGO\_RUBRIC\_GEO\_3D\_TILTED\_SOLID\_FILL}}

\begin{Verbatim}
card_id:        ALGO_RUBRIC_GEO_3D_TILTED_SOLID_FILL
shape:          algorithmic rubric
difficulty_tag: hard
domain_tags:    [geometry]
tier_visible:   HS
source:         meta_learner_4_24_epoch_1_rework
provenance:     supporting train problem P012; promotion_status = validated

routing_conditions (selection criteria):
  - vertex on plane
  - cube balanced
  - water level
  - horizontal plane
  - volume of water
  - tilted cube
  - body diagonal
routing_conditions (exclusion clauses):
  - solid is axis-aligned (not tilted)
  - problem asks about surface area rather than volume

payload (6-step rubric):
  1. Identify symmetry axis (body diagonal of the tilted solid).
  2. Compute vertex heights as dot products onto the vertical axis.
  3. Solve edge length first, before solving the fill-height question.
  4. Classify the water cross-section by which vertices fall below
     the fill height.
  5. Integrate, or decompose the submerged region into sub-pyramids
     and sum their volumes.
  6. Reduce the answer to p/q in lowest terms and output p + q.
\end{Verbatim}

This card was distilled for the ``tilted cube with water'' problem family that recurs on hard-geometry AIME items; it is shown verbatim, and its last step is answer formatting that the gate-judge did not remove. It lives at the (hard, geometry) cell of the store, so it is read into the active prompt whenever the test-time projection lands a problem at that cell --- that is, whenever the tiered pipeline escalates to HS and the coarse domain classifier (or its post-MS refinement) tags the problem as geometry.

\subsection*{Card 2: \texttt{MICRO\_GRID\_MAXIMAL\_ROW\_COL\_COLOR}}

\begin{Verbatim}
card_id:        MICRO_GRID_MAXIMAL_ROW_COL_COLOR
shape:          micro rubric
difficulty_tag: hard
domain_tags:    [combinatorics]
tier_visible:   HS
source:         train_add_4_25_v2_epoch_1
provenance:     supporting train problem P042; promotion_status = validated

routing_conditions (selection criteria):
  - n-by-n grid with chips
  - all chips in same row are same color
  - all chips in same column are same color
  - maximal placement, no additional chip can be placed
  - two-color row-column constraint
routing_conditions (exclusion clauses):
  - chip-counting problem without a maximality clause
  - single-color grids (no row/column color partition)

payload (4-step rubric):
  1. Classify rows and columns into states {W, B, E}
     (white-only, black-only, empty).
  2. Enforce color-agreement at every filled cell:
     W-row x W-col and B-row x B-col only.
  3. Enforce maximality: every empty cell must be FORCED by a row/col
     color clash; every E-row must intersect both W- and B-columns.
  4. Count admissible (#W-rows, #B-rows, #W-cols, #B-cols) tuples
     and multiply the corresponding binomials.
\end{Verbatim}

This card was distilled for the ``$n \times n$ grid, two-color chips, maximal placement'' problem family --- a recurring family in olympiad combinatorics. The card was distilled exclusively from training-split traces and was not constructed with reference to any test-set problem. It lives at the (hard, combinatorics) cell of the store. The four-step rubric encodes the maximality reduction that the empty-store hard solver consistently missed on training problems of this shape.

\paragraph{Reading these cards.} A card is read into the active prompt exactly when the test-time projection lands on its (difficulty, domain) cell: the difficulty tier is supplied by the tiered self-consistency pipeline (a card at \texttt{difficulty\_tag = hard} is reached only on escalation to HS), and the domain is supplied by the coarse classifier intersected with the card's \texttt{domain\_tags}. The card's content is its applicability claim; its position in the store is what determines whether that claim is consulted.

\section{Agent Prompts}
\label{app:agent-prompts}

This appendix reproduces verbatim the prompts used by the \textsc{KACE} offline learner (Algorithm~\ref{alg:kace}) and by the test-time pipeline. Each subsection is titled by the smallcaps function name as it appears in Algorithm~\ref{alg:kace}; trivial template tokens (problem text, candidate card lists, fire statistics) are denoted \texttt{<placeholder>}. The prompts below are drawn from our implementation.

\subsection{\textsc{BuildDomainPartition}}
\paragraph{Source.} \path{meta_learning/prompts/domain_tree.md}.
\begin{Verbatim}
SYSTEM:
You are the DOMAIN PARTITIONER for a competition-math reasoning system named
KACE. KACE has a knowledge bank of intervention cards organized by domain
(e.g. "geometry", "algebra"). At inference time the system classifies each
problem into a single domain and the solver only sees cards from that domain
(plus universal cards). Therefore the domain partition controls how much
irrelevant context the solver has to wade through -- too few coarse domains
= bloated card sets; too many narrow domains = solver never sees the right
card.

Your job: read a corpus of math problems and propose a disjoint partition
into 3-7 domains that minimizes context bloat per problem.

Partition rules:
1. Number of domains: between 3 and 7.
2. Each domain holds >=10% of the problems.
3. Disjoint by classification: a typical reader can assign each problem to
   exactly one domain based only on the problem statement.
4. Cluster by technique family, not surface topic.
5. Include a universal slot for cross-domain sanity checks (not counted
   toward the 3-7).

Output strict JSON:
{
  "rationale": "1-3 sentences",
  "domains": [
    {
      "name": "snake_case_short_name",
      "description": "1-line description",
      "size_pct": 25,
      "membership_signals": ["short phrase 1", "short phrase 2", ...]
    }
    /* ... 3-7 entries ... */
  ]
}

USER (template):
PROBLEM CORPUS (N problems sampled from training set):

[1] <problem 1 text>
EXPECTED: <answer 1>
[2] <problem 2 text>
EXPECTED: <answer 2>
... up to ~30 problems ...

Propose the partition as JSON (no prose outside JSON).
\end{Verbatim}

\subsection{\textsc{FindKnowledge}}
The Phase-1 per-cell sweep (\textsc{FindKnowledge}) performs both pattern search and card distillation in a single prompt; the orchestrator dispatches one call per learned-card cell.
\paragraph{Source.} \path{meta_learning/prompts/propose_per_domain.md}.
\begin{Verbatim}
SYSTEM:
You are the CARD PROPOSER for KACE. KACE has a 3-tier solver
(Easy / Medium / Hard) and a knowledge bank of intervention cards.
Each card is a short, reusable rule, formula, or rubric that the
solver reads at inference time to avoid a known failure mode.

Your job: read the wrong-attempt traces of problems in a single
domain and propose new cards that, if shown to the solver next time,
would prevent the first concrete mistake in those traces.

Card schema (strict):
{
  "card_id": "PREFIX_DOMAIN_TECHNIQUE_KEYWORD",
  "payload": "Useful when:\n- pattern1 (<=5 words)\n- pattern2\n- pattern3\nDo not apply if:\n- exclusion1 (<=5 words)\n\n<body>",
  "difficulty_tag": "medium" | "hard" | "universal",
  "domain_tags": ["<domain>"]
}

Difficulty tag rule:
- medium: short formula / identity / single-pass procedure (<=6 lines)
- hard: multi-step rubric (<=12 lines)
- universal: applies across all domains (use sparingly)

Card-ID style (ALL_CAPS_SNAKE_CASE, 3-6 tokens):
- EXACT_*       -- formula or identity
- RUBRIC_* / ALGO_RUBRIC_*  -- multi-step procedure
- READING_*     -- problem-misreading guard
- CASE_SPLIT_*  -- case-completeness reminder
- ANTI_*        -- meta-rule

Hard rules (self-reject any card that violates):
1. Answer leakage: never quote a numeric answer from any problem.
2. Instance overfit: replace problem-specific names with abstract analogs.
3. Benchmark neutrality: no mentions of "AIME", "USAMO", "[0,999]",
   "mod 1000", or any benchmark-specific format.
4. No vague exhortations.
5. Body length: medium <=6 non-blank lines; hard <=12; universal <=4.
6. Domain tag must match the input bucket's domain unless universal.

Volume guidance:
For each problem propose 1-3 cards. Default to 2.

Output strict JSON:
{ "domain": "<input domain>",
  "n_problems_addressed": <int>,
  "cards": [ /* array of cards */ ] }
Return ONLY this JSON.

USER (template):
DOMAIN: <domain>
EXISTING KB CARDS (read for format reference; do NOT duplicate):
[abbreviated list of card_ids and their first-line trigger]

PROBLEMS WITH WRONG ATTEMPTS (N entries):
[1] problem_id=P0XX
    PROBLEM: <text>
    EXPECTED: <answer>
    WRONG ATTEMPTS:
      - tier=MS, idx=0, wrong_ans=...
        excerpt: <last 1500 chars of trace>
      - ...
    EXIT: <es_unanimous|ms_majority|hs_plurality|...>
    CARDS SHOWN MS: [...]
    CARDS SHOWN HS: [...]
[2] ...

Propose new cards as JSON.
\end{Verbatim}

\subsection{\textsc{ReflectAndReviseCards}, \textsc{Refine}, and \textsc{GateJudge}}
The Phase-2 per-card revision (\textsc{ReflectAndReviseCards}), the \textsc{Refine} branch that self-corrects over-large edits, and the leak/quality/tier-coherence checks of \textsc{GateJudge} are folded into a single curator prompt; \textsc{GateJudge}'s constraints appear inside the strict-card-schema block reused from the proposer above (answer-leakage, instance-overfit, benchmark-neutrality), and \textsc{Refine} is operationalized as the EDIT branch of the curator's KEEP/EDIT/DEPRECATE decision.
\paragraph{Source.} \path{meta_learning/prompts/curate_per_domain.md}.
\begin{Verbatim}
SYSTEM:
You are the CURATOR for KACE's intervention cards in a single domain.
You will see (a) every card in this domain, (b) per-card fire statistics
from a recent train run, and (c) the training problems in this domain
that were still wrong after that run. Validation outcomes are NEVER
shown to you and never feed into the curation decision. Your job is to
clean up the bucket so the next run benefits from a tighter, more
targeted KB.

Decision options for each card:
- KEEP: card fired >=2 times AND shown_correct >= shown_wrong / 2.
- EDIT: card content sound but trigger header too vague or too narrow.
        Rewrite the trigger header (and optionally the body); each
        trigger bullet <=5 words; stay within tier line caps.
- DEPRECATE: drop the card entirely. Use when:
  - the card never fired and its content is too narrow to ever apply
    outside the original problem,
  - it is redundant with another card in the same domain
    (note which one in `reason`),
  - it consistently fires on wrong problems with
    shown_wrong > shown_correct + 1,
  - the body contains a wrong/misleading rule.

Decision priors (use to break ties):
- For cards with n_shown == 0: prefer EDIT if technique-rich;
  prefer DEPRECATE if problem-specific recipe.
- For cards with shown_correct >> shown_wrong: KEEP.
- For cards introduced in this epoch with n_shown == 0: prefer DEPRECATE.

No new cards in Phase 2:
You MUST NOT propose any new cards in this phase. Phase 2 is strictly
edit-only over the existing card set; new-card distillation happens in
Phase 1. If the wrong-problems list reveals a failure mode that no
existing card can be edited to cover, mark the affected cards
DEPRECATE and surface the gap in the `reason` field; the next Phase 1
will pick it up.

Output strict JSON:
{
  "domain": "<input domain>",
  "decisions": [
    {"card_id": "...", "action": "KEEP"},
    {"card_id": "...", "action": "EDIT",
     "new_payload": "...", "reason": "..."},
    {"card_id": "...", "action": "DEPRECATE", "reason": "..."}
  ]
}
Return ONLY this JSON.

USER (template):
DOMAIN: <domain>

EXISTING CARDS (N entries with fire stats):
[card_id]
  difficulty: <tag>, domains: [...]
  payload: <full payload>
  fire stats: n_shown=N, shown_correct=C, shown_wrong=W,
              tier_breakdown={MS:.., HS:..}
  shown_on_problems: [...]

WRONG PROBLEMS IN THIS DOMAIN (M entries):
[1] problem_id=...
    PROBLEM: <text>
    EXPECTED: <answer>
    FINAL: <model answer>
    EXIT: <path>
    CARDS_SHOWN_MS: [...]
    CARDS_SHOWN_HS: [...]
[2] ...

Decide each card and return JSON.
\end{Verbatim}

\subsection{Coarse Domain Classifier (test-time)}
\paragraph{Source.} \path{kace/agents/domain_classifier.py}, \texttt{\_build\_system\_prompt()}.
\begin{Verbatim}
You are a math problem domain classifier.

Read the problem and pick the SINGLE best primary domain from:
{lines}
- "mixed"  -- only if the problem genuinely combines two domains as primary

Use the EXACT domain name (snake_case where shown). Downstream card filters
match on these exact names; deviations exclude relevant cards.

Respond with strict JSON: {"primary": <one of the listed domain names>}.
\end{Verbatim}
The list \texttt{\{lines\}} is dynamically expanded from the domain registry produced by \textsc{BuildDomainPartition}.

\subsection{Easy Solver (ES)}
\paragraph{Source.} \path{kace/agents/easy_solver.py}.
\begin{Verbatim}
SYSTEM:
You are a careful competition-math solver.

LENS FOR THIS ATTEMPT: {lens_prompt}

End your response with the final answer enclosed in \boxed{...}.

USER:
PROBLEM:
{problem_text}

Now solve the problem carefully.

Lenses (DIPPER 2-lens rotation across the 2 ES attempts):
- domain_patterns: Look for the underlying mathematical domain (number
    theory, combinatorics, geometry, algebra) and recall canonical
    techniques for problems of that shape. Name the technique before
    using it.
- computation_chain: Focus on the algebraic/arithmetic chain. Derive
    symbolically, avoid numerical approximation, and verify each step
    before continuing.
\end{Verbatim}

\subsection{Medium Solver (MS)}
\paragraph{Source.} \path{kace/agents/medium_solver.py}.
\begin{Verbatim}
SYSTEM:
You are a careful competition-math solver. You receive a small set of
MEDIUM-tier guidance cards selected for this problem.

Use the cards where their premise fits. Quote any formula or rule
explicitly when you use it.

LENS FOR THIS ATTEMPT: {lens_prompt}

End your response with the final answer enclosed in \boxed{...}.

{card_block}

USER:
PROBLEM:
{problem_text}

Now solve the problem carefully, applying any relevant card.

Lenses (DIPPER 3-lens rotation across the 3 MS attempts):
- domain_patterns:    Look for the underlying mathematical domain and
                      recall canonical techniques.
- computation_chain:  Focus on the algebraic chain. Derive symbolically
                      and verify each step.
- problem_setup:      Read the problem precisely. State every condition
                      and constraint cleanly.
\end{Verbatim}

\subsection{Hard Solver (HS)}
\paragraph{Source.} \path{kace/agents/hard_solver.py}.
\begin{Verbatim}
SYSTEM:
You are a careful competition-math solver attacking a HARD problem.
Prior tiers (Easy, Medium) did not converge.

Apply the HARD-tier guidance cards where their premise fits. Quote any
rule or rubric step you follow. If a card is an algorithmic rubric,
walk through its steps.

End your response with the final answer enclosed in \boxed{...}.

{card_block}

USER:
PROBLEM:
{problem_text}

Now solve the problem carefully, applying the hard cards above.
\end{Verbatim}

\section{Implementation Details}
\label{app:impl}

This appendix records the exact per-tier configuration used for the main results (Table~\ref{tab:tiered-config}), the measured per-problem solver-call counts that justify the compute claims of §\ref{sec:tiered-vs-bo5} (Table~\ref{tab:calls-per-problem}), and the size and accuracy of the frozen card store per dataset (Table~\ref{tab:kb-size}).

\subsection{Per-Tier Configuration}

\begin{table}[t]
    \centering
    \small
    \begin{tabular}{lccc}
        \toprule
        Tier / Component & MATH-HARD & OlymMATH-EN & AIME25 \\
        \midrule
        model & gpt-4o-mini & gpt-4.1-mini & gpt-4.1-mini \\
        ES (2 attempts, $2/2$ unanimous) & $T{=}0.6$, $2$k tok  & $T{=}0.6$, $6$k tok  & $T{=}0.6$, $6$k tok  \\
        MS (3 attempts, $2/3$ majority)  & $T{=}0.6$, $2$k tok  & $T{=}0.6$, $12$k tok & $T{=}0.6$, $12$k tok \\
        HS (5 attempts, $2/5$ plurality) & $T{=}0.8$, $2$k tok  & $T{=}0.8$, $12$k tok & $T{=}0.8$, $12$k tok \\
        Coarse domain classifier         & $T{=}0$, $512$ tok  & (same) & (same) \\
        \bottomrule
    \end{tabular}
    \caption{Per-tier configuration used for all main-table runs. Output-token caps differ by dataset: the MATH-HARD run reuses a $2$k cap from an earlier sweep, while OlymMATH-EN and AIME25 use the default $6$k/$12$k/$12$k caps. The hard-solver temperature is $T{=}0.8$ across all three datasets to encourage sampling diversity at the tier where coverage rather than agreement is the binding constraint; ES and MS solver temperatures are $T{=}0.6$. The coarse domain classifier is deterministic.}
    \label{tab:tiered-config}
\end{table}

Table~\ref{tab:tiered-config} gives the exact knobs of the locked pipeline. The split between solver temperature ($T{=}0.6$ or $0.8$) and the deterministic decoding ($T{=}0$) of the coarse domain classifier reflects different roles: solvers benefit from sampling diversity inside a self-consistency vote, whereas the domain projection that selects the active cell of the store must reason deterministically so that the cards reaching the solver are reproducible across reruns.

\subsection{Solver Calls per Problem}

  \begin{table}[t]
      \centering
      \small
      \begin{tabular}{lccc}
          \toprule
          Method & MATH (/50) & OlymMATH-EN (/27) & AIME25 (/30) \\
          \midrule
          Best-of-$5$ self-consistency        & 5.00 & 5.00 & 5.00 \\
          Tiered SC (no cards) & 3.19 & 5.24 & 5.40 \\
          \textsc{KACE} (frozen store)          & 3.41 & 5.22 & 5.24 \\
          \bottomrule
      \end{tabular}
    \caption{Average solver calls per problem (ES + MS + HS only; routers and domain classifiers excluded). Tiered
  baseline budgets: ES = 2, MS = 3, HS = 5 (floor 2.0, ceiling 10.0). 3-seed averages where available.}
    \label{tab:calls-per-problem}
\end{table}

Table~\ref{tab:calls-per-problem} grounds the comparable-compute claim made in §\ref{sec:tiered-vs-bo5}. MATH-HARD is dominated by ES exits and averages well below the Best-of-$5$ budget; OlymMATH-EN and AIME 2025 escalate more frequently to MS and HS and average slightly above $5$ calls. In all cases the budget remains close to fixed Best-of-$5$ because no problem can exceed the $2{+}3{+}5 = 10$ budget and many exit early at ES or MS.

\subsection{Knowledge-Base Size per Dataset}

\begin{table}[t]
    \centering
    \small
  \begin{tabular}{lcccc}
      \toprule
      Dataset & Active cards & Total store tokens & Avg.\ tokens/card & Test accuracy (mean) \\
      \midrule
      MATH-HARD (50)   & 49  & $\sim$6{,}400  & 130 & 77.3\% \\
      OlymMATH-EN (27) & 40  & $\sim$4{,}500  & 113 & 48.1\% \\
      AIME25 (30)      & 132 & $\sim$33{,}600 & 254 & 61.1\% \\
      \bottomrule
  \end{tabular}
    \caption{Frozen card store per dataset (epoch count chosen on validation; the store is frozen before any test run). ``Total store tokens'' is the maximum context length if every card in $\mathcal{K}$ were shown to the solver simultaneously; in practice the test-time projection lands on a single (difficulty, domain) cell and only the cards under that cell reach the solver: $\sim$5{,}560 tokens per escalated attempt on the 132-card AIME store (measured, Table~\ref{tab:budget-comparison}), proportionally less on the smaller stores.}
    \label{tab:kb-size}
\end{table}

Table~\ref{tab:kb-size} reports the size of the frozen card store used at evaluation. The numbers make explicit a property that is central to the storage-vs-injection separation argument of §\ref{sec:formalism}: although the durable store carries thousands of tokens of distilled knowledge, only the slice under the active (difficulty, domain) cell --- $\sim$5{,}560 tokens per escalated attempt on the AIME store, roughly a sixth of the stored capacity (measured, Table~\ref{tab:budget-comparison}) --- reaches the solver on escalated MS/HS attempts. This is the operational form of decoupling capacity from active context size, and it is what allows \textsc{KACE} to grow the store without paying a proportional active-context tax. Test accuracy entries report the mean headline numbers in Table~\ref{tab:main-exp}; all three benchmarks are reported as $3$-seed averages.

\subsection{Compute Accounting}

All reported experiments use hosted API models; no local GPU training or weight updates are performed. Inference-time compute is reported in two ways: Table~\ref{tab:tiered-config} gives the per-tier output-token caps and Table~\ref{tab:calls-per-problem} reports measured average solver calls per problem on the test split. Our implementation reproduces the AIME 2025 result from the frozen 132-card store using \path{scripts/reproduce_paper.py}; its default runner uses API parallelism rather than local accelerator compute.

The offline construction loop consumes Claude Sonnet 4.6 calls through the Anthropic Agent SDK for domain construction, card proposal, and curation, and consumes OpenAI API calls when the train/validation runner evaluates candidate stores. Table~\ref{tab:offline_learner_cost} reports the Sonnet-side offline learner token budget and dollar cost for the two-epoch construction setting. Table~\ref{tab:solver_cost} reports the solver-side gpt-4.1-mini cost for the corresponding AIME run. The compute comparison supporting the main empirical claims should still be read as an inference-time solver-call comparison; the tables below account for the offline learner and solver costs that build and evaluate the released card bank.

\begin{table}[t]
    \centering
    \small
    \begin{tabular}{lccc}
        \toprule
        Bucket & Tokens & Rate (\$/MTok) & Cost \\
        \midrule
        Cache write   & 0.10M & 3.75  & \$0.38 \\
        Cache read    & 1.00M & 0.30  & \$0.30 \\
        Fresh input   & 0.20M & 3.00  & \$0.60 \\
        Output        & 0.32M & 15.00 & \$4.80 \\
        \midrule
        \textbf{Per epoch} & 1.62M & --- & \textbf{\$6.08} \\
        \bottomrule
    \end{tabular}

    \vspace{0.5em}
    \begin{tabular}{lc}
        \toprule
        Scenario (2 epochs, 132 AIME cards) & Cost \\
        \midrule
        Uncached            & \$17.40 \\
        Cached              & \$12.20 \\
        Aggressive cache    & \$8.00 \\
        \bottomrule
    \end{tabular}
    \caption{Offline learner cost on Sonnet 4.6 with Anthropic prompt caching: per-epoch input/output token split and 2-epoch totals for the 132-card AIME output. Cache write $=\$3.75$/MTok ($1.25\times$ base), cache read $=\$0.30$/MTok ($0.1\times$), base input $=\$3$/MTok, output $=\$15$/MTok.}
    \label{tab:offline_learner_cost}
\end{table}

\begin{table}[t]
    \centering
    \small
    \begin{tabular}{lccc}
        \toprule
        Phase & Problem-runs & Uncached & Cached \\
        \midrule
        Train ($\times 2$)         & 90  & \$1.71 & \$1.32 \\
        Val ($\times 2$)           & 90  & \$1.71 & \$1.32 \\
        Impact eval                & 90  & \$1.71 & \$1.32 \\
        Propose / PCS              & 60  & \$1.14 & \$0.88 \\
        Test ($\times 1$)          & 30  & \$0.57 & \$0.44 \\
        \midrule
        \textbf{Total}             & 360 & \textbf{\$6.84} & \textbf{\$5.28} \\
        \bottomrule
    \end{tabular}
    \caption{Solver-side cost on gpt-4.1-mini for a 2-epoch \textsc{KACE} run on AIME (train 45, validation 45, test 30; solver calls only, with routers and domain classifiers excluded; pricing $\$0.40$/$\$0.10$/$\$1.60$ per MTok input/cached/output).}
    \label{tab:solver_cost}
\end{table}

Solver-side cost is smaller than the cached offline learner cost on Sonnet 4.6 (Table~\ref{tab:offline_learner_cost}). The dominant solver-side driver is output tokens at the MS and HS tiers.

\subsection{Budget Comparison with Learned-Context Baselines}
\label{app:budget-comparison}

Matching test-time solver calls does not by itself establish that \textsc{KACE} and the learned-context baselines consume comparable overall resources. Table~\ref{tab:budget-comparison} therefore compares the three methods on AIME 2025 along the axes that differ: the offline optimization budget spent constructing the learned artifact, the teacher-side reflection budget, the capacity of the stored artifact, and the learned-context tokens actually injected at test time.

\begin{table}[t]
    \centering
    \footnotesize
    \setlength{\tabcolsep}{4pt}
    \begin{tabular}{lcccc}
        \toprule
         & Tiered & Tiered $+$ ACE & Tiered $+$ GEPA & \textsc{KACE} \\
        \midrule
        Offline problem-runs               & ---  & 45            & $\sim$280 (300 budget) & 360 \\
        Offline solver calls (est.)        & ---  & $\sim$250     & $\sim$1{,}540          & $\sim$1{,}870 \\
        Teacher (Sonnet 4.6) calls         & ---  & 17 reflections & 10 reflection batches & learner loop (Table~\ref{tab:offline_learner_cost}) \\
        Learner cost (metered)             & ---  & ---           & ---                    & \$12.20 cached \\
        Learned artifact                   & none & 17-bullet playbook & single HS prompt  & 132-card store \\
        Stored capacity (tokens)           & 0    & 814           & 532                    & $\sim$33{,}600 \\
        Learned tokens / escalated attempt & 0    & 814 (HS only) & 532 (HS only)          & $\sim$5{,}560 (MS$+$HS) \\
        Learned input tokens / problem     & 0    & ---           & 857                    & 18{,}064 \\
        Active input tokens / problem (total) & 1{,}498 & ---           & $\sim$2{,}000   & 19{,}561 \\
        Test-time solver calls / problem   & 5.40 & ---           & 5.38                   & 5.24 \\
        \bottomrule
    \end{tabular}
    \caption{Offline and test-time budget comparison on AIME 2025. ``Problem-runs'' counts full passes of a problem through the tiered pipeline during offline optimization; solver-call estimates multiply problem-runs by the measured $\sim$5.2--5.5 calls per run. ACE and GEPA teacher calls are counted from the archived reflection logs; their token totals were not metered, so no dollar figure is reported. \textsc{KACE}'s active-input row is \emph{measured} by re-tokenizing every stored solver prompt (system $+$ user, ES/MS/HS attempts only; routers and classifiers excluded) across all three seeds of the paired re-runs with the gpt-4.1 tokenizer. The baseline runners do not store prompts, so the Tiered $+$ GEPA figure is reconstructed from that arm's own prompt template, its measured call counts, and its learned-text size: its fixed 532-token prompt on each of its measured $1.61$ HS attempts per problem, plus its $\sim$210-token base prompt on each of $5.38$ calls. The Tiered $+$ ACE arm left no records we can match, neither per-problem outcomes nor stored prompts, so we do not estimate its per-problem token budget or call count.}
    \label{tab:budget-comparison}
\end{table}

Two asymmetries are worth stating plainly. First, matched solver calls do not mean matched tokens: measured over every stored prompt of the paired re-runs, \textsc{KACE} consumes $19{,}561$ input tokens per test problem against $1{,}498$ for the card-free scaffold and a reconstructed $\sim$$2{,}000$ for Tiered $+$ GEPA --- roughly $13\times$ the scaffold and about ten times Tiered $+$ GEPA, because the reference configuration renders the projected card pool (up to 30 cards) ($\sim$$5{,}560$ learned tokens per escalated MS/HS attempt --- $18{,}064$ learned tokens per problem over a measured $3.24$ MS$+$HS attempts per problem --- on the $60\%$ of problems that escalate). These counts cover solver input tokens only, so we make no total-API-cost comparison across arms; the storage-vs-usage claim should be read as capacity being decoupled from the active slice, not as the active slice being minimal. Second, \textsc{KACE}'s offline solver budget (360 problem-runs) is comparable to GEPA's $\sim$280 problem-runs (300-rollout budget), while ACE's single epoch is substantially cheaper; the additional \textsc{KACE}-specific cost is the Sonnet-side learner loop of Table~\ref{tab:offline_learner_cost}, which is about \$12 cached (\$17.40 uncached) for the full two-epoch construction. Under this accounting the comparison against GEPA is budget-matched on the offline solver side, and \textsc{KACE} carries a modest extra teacher cost and a $\sim$$13\times$ test-time input-token overhead, in exchange for a learned store roughly $60\times$ the capacity of the strongest baseline prompt.

\section{Paired Problem-Level Statistics on AIME 2025}
\label{app:paired-stats}

All AIME25 arms marked $^\dagger$ in Table~\ref{tab:main-exp} were re-run under a fully paired protocol on the same 30-problem test split, with per-problem outcomes recorded for every seed. Tiered $+$ ACE was not re-run and is excluded from the paired analysis.

\paragraph{Method.} The unit of analysis is the problem ($n{=}30$). For each arm we report per-seed correct counts and an exact Clopper--Pearson $95\%$ interval on the pooled proportion. For each comparison against \textsc{KACE} we compute the per-problem difference of seed-mean accuracies and report (i) a $95\%$ paired-bootstrap interval over problems ($20{,}000$ resamples) and (ii) a two-sided sign-flip permutation test on the per-problem differences, enumerated exactly over all $2^m$ sign assignments of the $m$ nonzero differences (no Monte-Carlo error). An exact McNemar test on seed-indexed run pairs is reported as a secondary check only: seed-index alignment across arms is arbitrary, and McNemar treats the $30 \times S$ paired trials as independent, which is anticonservative under within-problem seed correlation. A permutation $p$ of exactly $1.0$ arises when every sign assignment yields a mean difference at least as extreme as the one observed (as for \textsc{KACE} vs.\ Tiered, where all seven nonzero differences have magnitude $1/3$).

\begin{table}[t]
    \centering
    \small
    \begin{tabular}{lccc}
        \toprule
        Arm & Seeds (correct/30) & Pooled & $95\%$ CI \\
        \midrule
        \textsc{KACE} & 20, 19, 16 & $55/90 = 61.1\%$ & $[50.3\%,\ 71.2\%]$ \\
        Tiered SC (no cards) & 20, 17, 17 & $54/90 = 60.0\%$ & $[49.1\%,\ 70.2\%]$ \\
        Tiered $+$ GEPA & 16, 19, 17 & $52/90 = 57.8\%$ & $[46.9\%,\ 68.1\%]$ \\
        Best-of-$5$ & 16, 17, 15 & $48/90 = 53.3\%$ & $[42.5\%,\ 63.9\%]$ \\
        Best-of-$5$ $+$ ACE (minimal prompt) & 17, 16, 16 & $49/90 = 54.4\%$ & $[43.6\%,\ 65.0\%]$ \\
        \textsc{KACE} w/o difficulty axis & 18, 17, 17 & $52/90 = 57.8\%$ & $[46.9\%,\ 68.1\%]$ \\
        \bottomrule
    \end{tabular}
    \caption{Per-arm AIME25 results under the paired protocol: per-seed correct counts out of 30, pooled accuracy, and exact Clopper--Pearson $95\%$ intervals. All arms have three seeds.}
    \label{tab:paired-arms}
\end{table}

\begin{table}[t]
    \centering
    \small
    \setlength{\tabcolsep}{4pt}
    \begin{tabular}{lccccc}
        \toprule
        \textsc{KACE} vs. & $\Delta$ (pts) & $95\%$ bootstrap CI & Exact $p$ & B/W/T & McNemar (sec.) \\
        \midrule
        Best-of-$5$ & $+7.8$ & $[+0.0,\ +15.6]$ & $0.121$ & 7/2/21 & 11 vs.\ 4, $p{=}0.119$ \\
        Tiered SC (no cards) & $+1.1$ & $[-6.7,\ +8.9]$ & $1.000$ & 4/3/23 & 6 vs.\ 5, $p{=}1.000$ \\
        Tiered $+$ GEPA & $+3.3$ & $[-3.3,\ +11.1]$ & $0.563$ & 5/3/22 & 8 vs.\ 5, $p{=}0.581$ \\
        Best-of-$5$ $+$ ACE & $+6.7$ & $[+2.2,\ +11.1]$ & $0.031$ & 6/0/24 & 7 vs.\ 1, $p{=}0.070$ \\
        w/o difficulty axis & $+3.3$ & $[-5.6,\ +13.3]$ & $0.672$ & 4/4/22 & 9 vs.\ 6, $p{=}0.607$ \\
        \bottomrule
    \end{tabular}
    \caption{Paired problem-level comparisons, \textsc{KACE} vs.\ each arm on the same 30 problems. $\Delta$ is the mean per-problem difference of seed-mean accuracies (percentage points); the interval is a $95\%$ paired bootstrap over problems; $p$ is the exact two-sided sign-flip permutation test; B/W/T counts problems on which \textsc{KACE}'s seed-mean is better/worse/tied. McNemar (discordant counts $b$ vs.\ $c$, exact $p$) is secondary.}
    \label{tab:paired-comparisons}
\end{table}

\paragraph{Reading.} Every \textsc{KACE} margin on AIME25 is positive and none is conclusive at $n{=}30$. The smallest $p$-value, $0.031$ vs.\ Best-of-$5$ $+$ ACE ($+6.7$ points, better on 6 problems and worse on none), is one of five uncorrected comparisons, and we do not read it as establishing an effect: a 30-problem test cannot resolve differences of a few problems, and the observed margins ($+1.1$ to $+7.8$ points) sit below the minimum detectable effect at this size. The same exact paired machinery applied to MATH-HARD ($n{=}50$, three seeds per arm) gives \textsc{KACE} $+4.7$ points over Best-of-$5$ ($77.3\%$ vs.\ $72.7\%$; better on 6 problems, worse on 1, tied on 43, $p{=}0.109$). We therefore claim the ordering --- \textsc{KACE} above every comparison arm on AIME 2025 and above Best-of-$5$ on MATH-HARD --- and no statistical significance.

\section{Tier-Exit Concordance with Empirical Difficulty}
\label{app:tier-classifier}

Headline numbers appear in Table~\ref{tab:tier-classifier} (§\ref{sec:tier-classifier}); this appendix gives the per-bucket breakdown and discusses the dominant failure mode. This appendix evaluates how well the locked ES/MS/HS tier-exit ordering preserves the model's own empirical difficulty ordering of problems. The evaluation is purely model-derived: difficulty is defined by per-problem solve rate under independent sampling, with no human-supplied difficulty labels (e.g., AMC/AIME problem position, MATH curriculum level) entering the comparison.

\paragraph{Empirical difficulty.} For every test problem $i$ we run the locked tiered pipeline across $S=3$ independent random seeds; for MATH-HARD this uses the frozen MATH-HARD store. Each tier of the locked pipeline produces independent attempts (ES~$=2$, MS~$=3$, HS~$=5$); problems that exit early have fewer total attempts than those that escalate. Let $T_i$ be the total number of attempts pooled across the $S$ seeds (ranging from $6$ to $30$) and $C_i$ the number that produced the correct answer. The empirical difficulty of problem $i$ is
\begin{equation*}
    \mathrm{raw\_rate}_i \;=\; C_i / T_i \;\in\; [0, 1],
\end{equation*}
the fraction of independent samples that answer the problem correctly.

\paragraph{Predictor.} For each (problem, seed) we map the gate's exit path to an ordinal exit rank: higher rank means the gate \emph{predicts} an easier problem.

\begin{table}[ht]
    \centering
    \small
    \begin{tabular}{lcl}
        \toprule
        Exit path & Rank & Predicted difficulty \\
        \midrule
        \texttt{es\_unanimous}       & 4 & EASY      \\
        \texttt{ms\_majority}        & 3 & MEDIUM    \\
        \texttt{hs\_plurality}       & 2 & MED-HARD  \\
        \texttt{fallback\_plurality} & 1 & HARD      \\
        \texttt{fallback\_last\_hs}  & 0 & HARDEST   \\
        \bottomrule
    \end{tabular}
    \caption{Ordinal exit-rank predictor used in the concordance metric below.}
    \label{tab:exit-rank}
\end{table}

For each problem the exit-rank predictor is $e_i$, the mean rank across the $S$ seeds.

 \paragraph{Concordance metric.} We report a pairwise concordance score (equivalent to AUROC on the binary ``easier-than'' lift).
  Let $r_i = \mathrm{raw\_rate}_i$ and $e_i$ the predicted lift, and write $\Delta r_{ij} = r_i - r_j$, $\Delta e_{ij} = e_i - e_j$.
  Over ordered pairs $i<j$ with $\Delta r_{ij} \ne 0$:
  \begin{equation*}
      \rho_{\text{pair}} \;=\; \frac{\displaystyle\sum_{\substack{i<j \\ \Delta r_{ij} \ne 0}} \Big[\mathbf{1}\{\Delta r_{ij}\,\Delta
   e_{ij} > 0\} \;+\; \tfrac{1}{2}\,\mathbf{1}\{\Delta e_{ij} = 0\}\Big]}{\#\{i<j : \Delta r_{ij} \ne 0\}}.
  \end{equation*}
  A value of $1.0$ indicates perfect ordinal preservation, $0.5$ is the random baseline, and $0.0$ is fully inverted. We also report
  Spearman's $\rho$ on the pairs $(r_i, e_i)$ for completeness.

\begin{table}[h]
    \centering
    \footnotesize
    \setlength{\tabcolsep}{4pt}
    \begin{tabular}{lcccccccc}
        \toprule
        Difficulty bucket (raw\_rate) & \# probs & cells & ES & MS & HS & FB-plur & FB-last-hs \\
        \midrule
        EASY      ($\ge 80\%$)         & 8 & 24 & 23 & 1  & 0  & 0 & 0 \\
        MED-EASY  ($50$--$80\%$)        & 6 & 18 & 9  & 7  & 2  & 0 & 0 \\
        MED-HARD  ($20$--$50\%$)        & 6 & 18 & 0  & 4  & 12 & 2 & 0 \\
        HARD      ($1$--$20\%$)         & 2 & 6  & 1  & 0  & 3  & 1 & 1 \\
        VERY-HARD ($0\%$)               & 8 & 24 & 5  & 10 & 2  & 3 & 4 \\
        \bottomrule
    \end{tabular}
    \caption{AIME 2025 tier-exit cells per difficulty bucket. ``cells'' is the per-bucket total of (problem, seed) pairs ($\#\text{probs} \times S$). \textbf{Spearman's $\rho = 0.733$. Pairwise concordance $\rho_{\text{pair}} = 302/386 = 78\%$.}}
    \label{tab:concordance-aime}
\end{table}

\begin{table}[h]
    \centering
    \footnotesize
    \setlength{\tabcolsep}{4pt}
    \begin{tabular}{lcccccccc}
        \toprule
        Difficulty bucket (raw\_rate) & \# probs & cells & ES & MS & HS & FB-plur & FB-last-hs \\
        \midrule
        EASY      ($\ge 80\%$)         & 27 & 81 & 74 & 7  & 0 & 0 & 0 \\
        MED-EASY  ($50$--$80\%$)        & 10 & 30 & 14 & 14 & 2 & 0 & 0 \\
        MED-HARD  ($20$--$50\%$)        & 3  & 9  & 0  & 5  & 2 & 2 & 0 \\
        HARD      ($1$--$20\%$)         & 3  & 9  & 3  & 3  & 3 & 0 & 0 \\
        VERY-HARD ($0\%$)               & 7  & 21 & 7  & 12 & 1 & 1 & 0 \\
        \bottomrule
    \end{tabular}
    \caption{MATH-HARD tier-exit cells per difficulty bucket. \textbf{Spearman's $\rho = 0.816$. Pairwise concordance $\rho_{\text{pair}} = 743/923 = 80\%$.}}
    \label{tab:concordance-math}
\end{table}

\paragraph{Reading.} \textsc{KACE}'s tier-exit hierarchy preserves the empirical difficulty ordering of $78\%$ of AIME pairs and $80\%$ of MATH-HARD pairs --- substantially above the $50\%$ random baseline. The dominant failure mode on both benchmarks is the VERY-HARD bucket ($\mathrm{raw\_rate} = 0\%$): problems that the model never solves but on which the gate \emph{convergently exits early}, because all attempts agree on a wrong answer. On AIME 2025, $5+10 = 15$ of $24$ VERY-HARD cells exit at ES or MS; on MATH-HARD, $7+12 = 19$ of $21$ VERY-HARD cells do the same. These are the cases where the gate has no signal to detect difficulty --- agreement on a wrong answer is indistinguishable from agreement on a right answer at the level of self-consistency --- and they account for nearly all the discordant pairs. The same pathology surfaces in full \textsc{KACE} as wrongly selected cards on overconfident wrong answers, which is why calibration remains diagnostic rather than a closed loop. Without a step-level verifier or a learned judge, agreement is the strongest signal available, and agreement underestimates hardness in exactly the regime where harder reasoning is most needed.

\section{Single-Call Diagnostic: One Learned Artifact}
\label{app:diagnostic}

Before building the tiered schedule, we compared learned-context methods with one solver attempt per problem on AIME 2025 (\texttt{gpt-4.1-mini}, no self-consistency; Table~\ref{tab:aime25-prelim}). \emph{Domain-ACE} is ACE with a separate playbook per domain, each grown only from failures in that domain.

\begin{table}[ht]
    \centering
    \small
    \begin{tabular}{lc}
        \toprule
        Method & AIME25 \\
        \midrule
        Single-call baseline & 42.2\% \\
        ACE (single playbook, $\ge 2$ epochs) & 38.9\% \\
        GEPA (compact prompt, single epoch) & 51.1\% \\
        \textbf{Domain-ACE} (2 epochs) & \textbf{52.2\%} \\
        \bottomrule
    \end{tabular}
    \caption{Single-call diagnostic on AIME 2025. A single ACE playbook falls below the unaugmented solver; per-domain playbooks recover accuracy; GEPA's compact prompt avoids the drop.}
    \label{tab:aime25-prelim}
\end{table}

A single ACE playbook accumulated over two or more epochs scores below the unaugmented solver ($38.9\%$ against $42.2\%$). In its failures, irrelevant playbook entries sit next to the useful ones, and the solver follows entries that resemble the problem on the surface but do not apply to it. GEPA's compact prompt avoids this drop ($51.1\%$), but everything it learns has to fit in that one prompt. Partitioning the playbook by domain recovers and improves on the baseline after two epochs ($52.2\%$). With further epochs the per-domain playbooks grow as well: within a domain, rubrics for hard problems crowd out the entries that easy problems need, and accuracy falls back toward the single-playbook trajectory. Partitioning by domain addresses topic relevance but not difficulty relevance, which is why the \textsc{KACE} store adds the tier axis.

\section{Synthesis Experiments}
\label{app:synthesis}

We investigated whether \emph{context folding} across hard-solver attempts could be used to lift the hard tier without spending extra solver calls. The procedure was: after the HS pool of five attempts disagreed below the $2/5$ exit gate, the teacher LLM read the five partial trajectories, extracted a \emph{stable common prefix} of intermediate facts that the attempts agreed on (the ``stable point''), summarized that prefix into a compact context, and asked the solver to finish the solution from that synthesized state. The intent was to recover a $6$th attempt whose initialization conditioned on the agreed structure of the previous five rather than starting from the problem alone.

The empirical result was that synthesis produced essentially no change in accuracy, and the small movement that did appear was dominated by where the stable-prefix cut was placed. Cutting too early forfeited useful intermediate work that the attempts had already agreed on; cutting too late propagated whichever line of attack had drifted off-course in the majority of attempts, and the synthesized completion inherited the same error. Crucially, there was no reliable way to ascertain how much of each HS attempt's intermediate reasoning was actually correct: agreement across attempts is necessary but not sufficient for correctness, and without a step-level verifier the teacher could only summarize what \emph{looked} stable, not what was \emph{known} stable. This is the same bottleneck that limits self-refinement methods more generally: a refinement step is only as good as the refiner's ability to localize the actual point of divergence, and on hard math reasoning that localization is itself the unsolved subproblem. We therefore disable synthesis in the main pipeline and report it here only to record the failure mode.

\section{Imagination Experiments}
\label{app:imagination}

We also experimented with expanding the card store through \emph{imagination} --- generating cards that were not directly grounded in observed training failures, but were proposed by the teacher LLM as plausibly missing knowledge. Concretely, for each existing card $c$, the teacher was prompted along the following lines: \emph{``Suppose a base LLM is already using the errors and sanity-checks captured by this card but still gets a problem in this domain wrong. What else might it not know that an expert would consider obvious?''} The teacher proposed neighbor cards under this prompt, which were then put through the standard gate-judge admission and consolidation steps. Iterating this procedure expanded the store from the curated 132-card AIME store used in the main results to 450 cards.

\begin{table}[t]
    \centering
    \small
    \begin{tabular}{lccc}
        \toprule
        store variant & \#cards & AIME25 val & AIME25 test \\
        \midrule
        Curated only (cards from training-trace reflection) & 132 & 60.0\% & 61.1\% \\
        $+$ imagination, refinement round 1                 & 285       & 66.7\% & 62.2\% \\
        $+$ imagination, refinement round 2                 & 380       & 71.1\% & 61.1\% \\
        $+$ imagination, refinement round 3                 & 450       & 73.3\% & 60.0\% \\
        \bottomrule
    \end{tabular}
    \caption{Imagination expansion of the store. Imagined cards unlocked new solves on the validation split, but additional refinement rounds drove validation accuracy up while test accuracy plateaued and then drifted down --- the canonical signature of overfitting to a small validation set.}
    \label{tab:imagination}
\end{table}

The expanded store unlocked new solves on the validation split: problems that the curated store could not solve found a useful imagined card and were solved. However, two structural issues surfaced (Table~\ref{tab:imagination}). First, the \emph{quality} of imagined cards became very hard to check: gate-judging filters for ground-truth leakage and surface quality, but neither check distinguishes a genuinely missing piece of expert knowledge from a plausible-sounding restatement of something the model could already produce. Without a downstream verifier strong enough to test whether an imagined card encodes content the solver \emph{actually} lacks, lift on the validation split is the only available signal. Second, this validation-only signal was quickly exhausted: after a small number of refinement rounds the imagination loop had effectively memorized the validation residual, and further rounds raised validation accuracy without raising test accuracy --- the canonical signature of overfitting to a small evaluation set.

We therefore did not include imagined cards in the main \textsc{KACE} results: \textsc{KACE} defines its store as cards distilled from observed training failures (§\ref{sec:loop}), and imagined cards fall outside that definition even when they raise validation accuracy. The episode is informative on its own: it suggests that scaling the store beyond what training-trace failures can ground requires a verifier that operates above the level of paired train-verify lift, and that without such a verifier additional store capacity is not free --- it converts straightforwardly into validation overfitting. We treat this as a precursor to a judge-in-the-loop extension.

\end{document}